\documentclass[10pt,twocolumn,letterpaper]{article}

\usepackage[accsupp]{axessibility}  % Improves PDF readability for those with disabilities.
\usepackage{cvpr}              % To produce the CAMERA-READY version
\usepackage{algorithm}
\usepackage{algpseudocode}
\usepackage{amsmath}
\usepackage{amsmath, amssymb} 
\usepackage{enumitem}        
\usepackage[section]{placeins}  
\definecolor{cvprblue}{rgb}{0.21,0.49,0.74}
\usepackage[pagebackref,breaklinks,colorlinks,allcolors=cvprblue]{hyperref}

\usepackage{multirow}
\usepackage[table]{xcolor}

\def\cc{\cellcolor[RGB]{230,230,230}} % this color seems to be  more consistent with the style in introduction
\usepackage{pifont}
\newcommand{\cmark}{\ding{51}}%
\newcommand{\xmark}{\ding{55}}%
\newcommand{\snow}{\text{\ding{100}}} % Snowflake-like symbol
\newcommand{\frozen}{_{\text{\snow}}}
\newcommand{\tablestyle}[2]{\setlength{\tabcolsep}{#1}\renewcommand{\arraystretch}{#2}\centering\footnotesize}

\def\paperID{1211} % *** Enter the Paper ID here
\def\confName{CVPR}
\def\confYear{2026}

\title{DINO Eats CLIP: Adapting Beyond Knowns for Open-set
3D Object Retrieval}

\author{%
    Xinwei He$^{1}$
  ~~Yansong Zheng$^{1}$
  ~~Qianru Han$^1$
  ~~Zhichuan Wang$^1$
  ~~Yuxuan Cai$^2$
  ~~Yang Zhou$^3$ \\
  ~~Jingbo Xia$^1$
  ~~Yulong Wang$^1$
  ~~Jinhai Xiang$^1$ 
  ~~Xiang Bai$^2$\thanks{Correspondence to Xiang Bai$<$\href{xbai@hust.edu.cn}{xbai@hust.edu.cn}$>$.}\\
  $^1$Huazhong Agricultural University~~ $^2$Huazhong University of Science and Technology ~~ \\
  $^3$Shenzhen University\\
  {\footnotesize \texttt{xwhe@mail.hzau.edu.cn}, \texttt{xbai@hust.edu.cn}}
}

\begin{document}
\maketitle

\begin{abstract}
Vision foundation models have shown great promise for open-set 3D object retrieval (3DOR) through efficient adaptation to multi-view images.
Leveraging semantically aligned latent space, previous work typically adapts the CLIP encoder to build view-based 3D descriptors. Despite CLIP's strong generalization ability, its lack of fine-grainedness prompted us to explore the potential of a more recent self-supervised encoder—DINO. 
To address this, we propose DINO Eats CLIP (DEC), a novel framework for dynamic multi-view integration that is regularized by synthesizing data for unseen classes. 
We first find that simply mean-pooling over view features from a frozen DINO backbone gives decent performance. Yet, further adaptation causes severe overfitting on average view patterns of known classes.
To combat it, we then design a module named Chunking and Adapting Module (CAM). It segments multi-view images into chunks and dynamically integrates local view relations, yielding more robust features than the standard pooling strategy. 
Finally, we propose Virtual Feature Synthesis (VFS) module to mitigate bias towards known categories explicitly. 
Under the hood, VFS leverages CLIP's broad, pre-aligned vision-language space to synthesize virtual features for unseen classes. By exposing DEC to these virtual features, we greatly enhance its open-set discrimination capacity. 
Extensive experiments on standard open-set 3DOR benchmarks demonstrate its superior efficacy.
\end{abstract}    
\section{Introduction}
\label{sec:intro}

\begin{figure}[t]
\centering
\includegraphics[width=\linewidth]{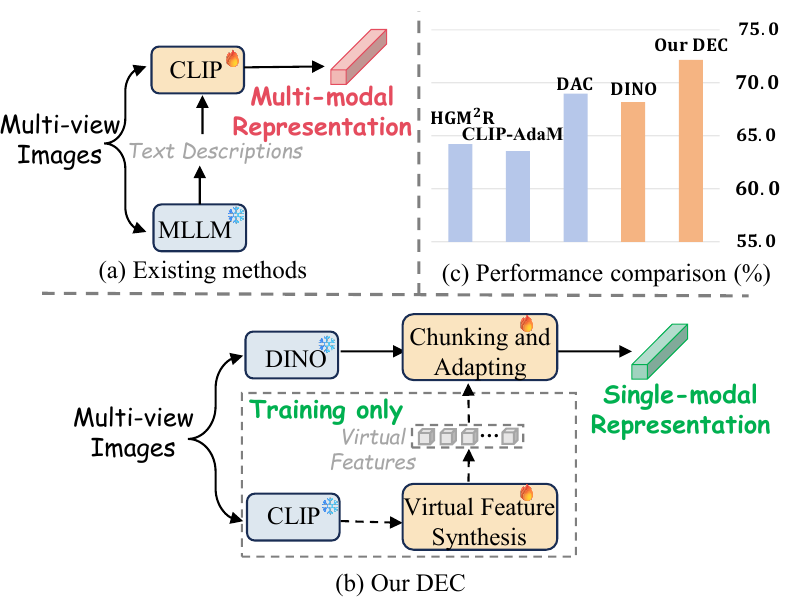}
\caption{(a) Prior CLIP-based methods rely on multi-modal image-text representations at inference time while our DEC (b) uses only single-modal visual features, making it more efficient for open-set 3DOR task. (c) Our DEC significantly outperforms existing methods on open-set 3D object retrieval. }
\label{fig:into_fig}
\vspace{-15pt}
\end{figure}

3D object retrieval (3DOR) has been extensively studied for many years~\cite{feng2018gvcnn,wei2020view, sitzmann2019deepvoxels, achlioptas2018learning,liu2019point2sequence, nie2019mmjn, jing2021cross, wu2019multi, he2024latformer}. Traditional methods generally assume a closed-set setting, where training and testing sets share the same label space. However, real-world applications demand generalized representations for unseen categories.
It has spurred a lately shift towards open-set learning, also known as open-set 3DOR~\cite{feng2022shrec}, which learns from a collection of 3D objects solely from seen categories but must generalize well to unseen categories.
 
Open-set 3DOR is a challenging task.
Limited 3D data from a few categories often leads to severe overfitting on seen classes and poor generalization to unseen ones.
The seminal work HGM$^2$R~\cite{feng2023hypergraph} integrates multi-modalities (point clouds, voxels, view images) of 3D objects to enrich the data, yet it requires a complex framework of modality-specific backbones, limiting its scalability and efficiency.
%Lately, recent 
Recent studies~\cite{he2025clip,wang2025describe} have demonstrated the efficacy of adapting pretrained vision foundation models like CLIP to multi-view images for this task (Figure~\ref{fig:into_fig}(a)). 
%However, they predominantly utilize text-aligned view embeddings from CLIP, leaving the potential of a more recent self-supervised encoder DINO untapped. 
Although the text-aligned view embedding from CLIP provides strong generalization capabilities, the limitations in fine granularity restrict its upper bound, prompting us to turn to a more recent self-supervised encoder---DINO.
We argue that DINO's inherent capacity to capture both local details and global structures in view images makes it a powerful source of robust and generalized view features for adaptation.

We first establish a simple multi-view DINO baseline. By simply mean-pooling across view embeddings, it provides a stronger zero-shot baseline than CLIP-based ones, as shown in Table~\ref{tab:main_results}.
However, further adaptation leads to severe overfitting to seen-class view features.
To mitigate it, we propose DINO Eats CLIP (DEC) (Figure~\ref{fig:into_fig}(b)), a novel framework that dynamically adapts a pretrained DINO encoder to multi-view images and uses CLIP as a bridge to achieve generalization to unseen classes. 
In contrast to view-pooling strategies that risk overfitting to pronounced view features, we introduce a Chunking and Adapting Module (CAM). This module avoids direct pooling of multi-view features. Instead, it first chunks multi-view embeddings into small groups and then aggregates each group into chunk features, capturing local inter-view patterns. Next, it combines cross-chunk features to capture longer-range inter-view relations. 
In this divide-and-conquer way, CAM gradually mines rich, multi-granular view patterns to produce a compact and generalized view-based 3D embedding.

Without exposure to unseen-class data, however, CAM still faces the risk of overfitting of known categories. 
One straightforward strategy is to synthesize virtual features of unseen classes for regularization, but this approach is hindered by the fundamental difficulty of learning from limited data.
Being contrastively trained on substantially broad concepts and images, CLIP~\cite{radford2021learning} provides a potential link between vision data of seen and unseen classes. 
We theoretically demonstrate that the semantic relations between classes are reflected in the structure of their visual spaces. 
Motivated by this theoretical insight, we further introduce a virtual feature synthesizer that transfers seen-unseen semantic relations in the text embedding space to the visual embedding space. 
By treating visual features of seen classes as anchors, we synthesize virtual visual features based on transferred semantic differences.
We finally train our CAM adapter to distinguish these virtual visual features of distinct unseen classes through an end-to-end metric learning loss, which effectively regularize CAM and further enhance open-set 3D object discrimination greatly.

In summary, we make the following contributions:
\begin{itemize}
    \item To our best knowledge, DEC is the first effort to adapt pretrained DINO for open-set 3D object retrieval. Equipping it with robust generalization to unseen categories from CLIP, we thereby fully leverage the strengths of this powerful self-supervised model to provide a concise yet effective solution.
    \item We introduce a lightweight Chunking and Adapting Module (CAM) that progressively aggregates multi-view features in a divide-and-conquer manner, capturing rich inter-view relations at different granularities for a generalized and compact 3D embedding.
    \item We develop a CLIP-based virtual feature synthesizer for regularization. Grounded by our theoretical insight that semantic relations are reflected in visual embedding space, it generates virtual unseen-class features by transferring textual semantic relations, enabling effective model regularization.
   
\end{itemize}

\section{Related Work}
\label{sec:related}

\noindent\textbf{3D Object Retrieval.}
3D Object Retrieval (3DOR) aims to learn discriminative representations from 3D objects. Existing approaches can be broadly divided into two categories: methods~\cite{maturana2015voxnet,wu20153d,wang2017cnn, qi2017pointnet, qi2017pointnet++, wang2019dynamic, liu2019relation, zhao2019pointweb} that learn directly from raw 3D data and methods~\cite{kanezaki2018rotationnet,han20193d2seqviews, esteves2019equivariant,feng2018gvcnn,dai2018siamese,he2018triplet, li2019angular,su2015multi,he2019view} that learn from 2D projections or views.
Although 3DOR has achieved remarkable progress, most existing methods assume that all test categories are seen during training, which limits their generalization to real-world applications. To address this limitation, \emph{open-set 3DOR}~\cite{feng2023hypergraph,feng2022shrec} has drawn increasing attention, where test 3D objects belong to unseen categories. This new setting evaluates the model’s ability to generalize beyond the training distribution, making it more suitable for practical deployment. 
Recent studies~\cite{zhou2023uni3d,xue2023ulip,xue2024ulip,wang2025teda,wang2025describe,he2025clip,song2023mv} have also explored this problem by using large-scale point cloud models or vision-language frameworks such as CLIP~\cite{radford2021learning}, aiming to learn generalized 3D representations through semantic alignment. 

While most existing methods are built upon CLIP, its global image-text alignment strategy limits the ability to capture fine-grained visual differences. In contrast, the recent self-supervised model DINO~\cite{caron2021dinov1,oquab2024dinov2,simeoni2025dinov3}  exhibits strong fine-grained representation extraction capacity and has demonstrated superior transfer ability in a variety of computer vision tasks, such as segmentation~\cite{barsellotti2025talking,jose2025dinov2}, 3D reconstruction~\cite{wang2025vggt}, and industrial anomaly detection~\cite{guo2025dinomaly}. 
Yet, research on adapting it for 3D representation learning is underexplored. 
Therefore, in this paper, we explore the potential of DINO for open-set 3DOR. Instead of utilizing simple multi-view feature pooling~\cite{he2025clip,wang2025describe} which ignores inter-view relations, we propose a Chunking and Adapting Module (CAM) to integrate rich local inter-view patterns for discriminative and generalized view-based descriptors.

\noindent\textbf{Virtual Feature Exposure.}
Our work is also relevant to Out-of-distribution (OOD) detection~\cite{yang2024generalized}.
To mitigate lack of knowledge from unknowns, one effective strategy~\cite{duvos} is to synthesize virtual features for \emph{regularization}, which helps to learn a compact decision boundary for OOD detection. 
However, they generally treats all conflate all unseen classes into a monolithic “unknown” one, which is infeasible for Open-set 3DOR, which requires explicit associations between samples and their semantic categories to improve open-set discriminative learning.
Recently, some works, e.g., APLGOS~\cite{zhang2025adaptive} and SHIP~\cite{wang2023improving}, use a pretrained CLIP text encoder to synthesize virtual prompts by training a generative model (\emph{e.g.}, variational autoencoder) to enhance downstream 2D vision tasks. 
Yet, it is difficult for 3D domain due to data scarcity.
In contrast, we directly leverage the text-vision alignment of vision-language models (CLIP) to connect unseen and seen categories for regularization. To this end, we synthesize semantically grounded unseen features to effectively enhance open-set 3DOR.

\section{Method}
\label{sec:method}

\begin{figure*}[ht]
\centering
%% old: pipeline_v2
\includegraphics[width=0.95\linewidth]{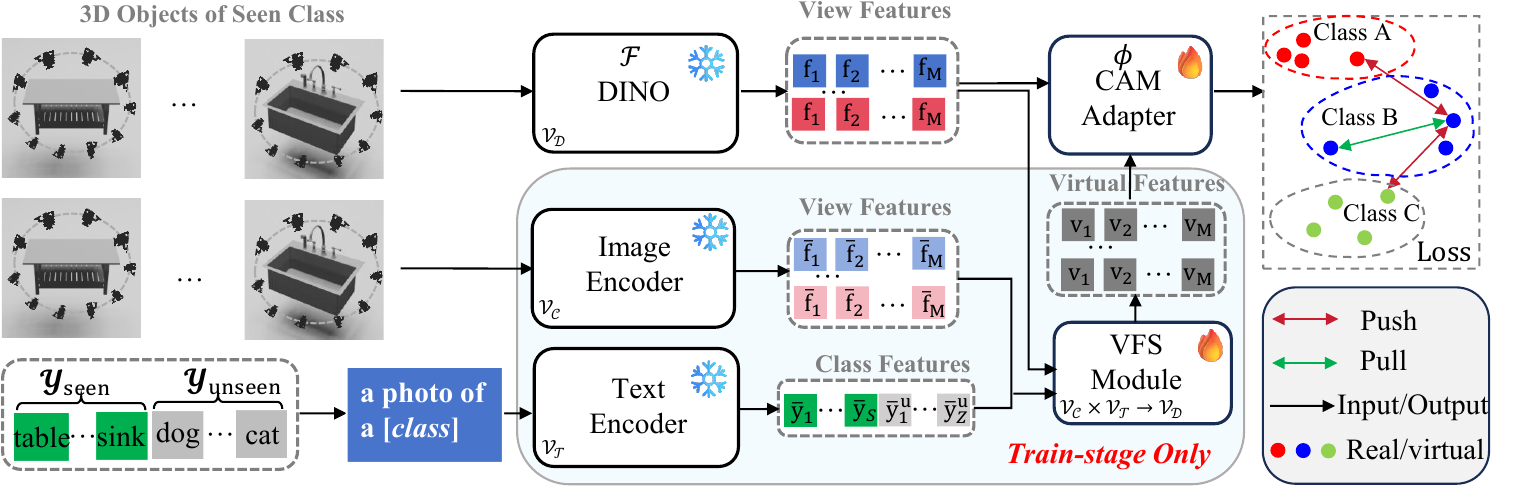}
\caption{\textbf{Overview of our framework}. During training, it processes known-category images by frozen DINO and CLIP image encoders, while encoding known and unseen (\eg, ImageNet) class labels via CLIP's text encoder using the prompt "a photo of [class]". Based on them, we synthesize virtual features to train our chunking and adapting (CAM) adapter jointly with an end-to-end metric learning loss, encouraging it to produce generalized view-based 3D descriptors to unseen categories.}
\label{fig:framework}
\vspace{-8pt}
\end{figure*}

\subsection{Problem Statement and Overview} 
\noindent\textbf{Problem statement}. Open-set 3DOR aims to learn an embedding network $f_{\mathbf{\Theta}}(\cdot)$ that induces a metric space where 3D objects of distinct unseen categories form compact, well-separated clusters.
Formally, we are given a collection $\mathcal{O}$ of 3D objects and a training set with $N_{tra}$ samples from a closed label set $\mathcal{Y_{\text{seen}}}$ to train $f_{\mathbf{\Theta}}(\cdot)$,  \emph{i.e.}, $\mathcal{D}_{\text{tra}} = \{(\mathbf{o}^i, y^i)\}_{i=1}^{N_{tra}}$, where $\mathbf{o}^i \in \mathcal{O}$ denotes the $i$-th 3D object and $y^i \in \mathcal{Y}_{\text{seen}}$ represents its label.
At test time, we will apply $f_{\mathbf{\Theta}}(\cdot)$ to embed a retrieval set with $N_{ret}$ samples from unseen category space $\mathcal{Y_{\text{unseen}}}$, \emph{i.e.}, $\mathcal{D}_{\text{ret}}=\{(\mathbf{o}^j, y^j)\}_{j = 1}^{N_{ret}}$, where $\mathbf{o}^j \in \mathcal{O}$ and $y^j \in \mathcal{Y}_{\text{unseen}}$.
Note that $\mathcal{Y_{\text{seen}}}$ and  $\mathcal{Y_{\text{unseen}}}$ are non-overlapping label spaces indicating $\mathcal{Y_{\text{seen}}} \cap \mathcal{Y_{\text{unseen}}}=\emptyset$. 
$\mathcal{D}_{\text{ret}}$ is further divided into query set $\mathcal{D}_q$ and target set $\mathcal{D}_t$. 
Each 3D object $\mathbf{o}^j$ from $\mathcal{D}_q$ serves as the query to retrieve from $\mathcal{D}_t$ to evaluate the generalizalibity of $f_{\mathbf{\Theta}}(\cdot)$. 
% achieving both high intra-class cohesion and inter-class separation for novel classes.

\noindent\textbf{Overview}. Figure~\ref{fig:framework} presents an overview of our framework. In essence, it encodes multi-view images with frozen, pretrained models (denoted by $\mathcal{F}_{\frozen}(\cdot)$), and then adapts to these multi-view embeddings with a learnable adapter (denoted by $\mathcal{\phi}(\cdot)$), \emph{i.e.},
$f_{\mathbf{\Theta}}(\cdot) = (\mathcal{\phi} \circ \mathcal{F}_{\frozen}) (\cdot)$.
The adapter divides the multi-view images into small chunks and learn to integrate rich local patterns across views, producing discriminative and generalized representations. 
To avoid overfitting on known categories, we further build upon the CLIP model and design a novel virtual feature synthesis (VFS) module to synthesize virtual visual features to regularize the training process. 
By design, it simply integrates three feature sources (the DINO embedding space $\mathcal{V_D}$, the CLIP visual embedding space $\mathcal{V_C}$ and text embedding space $\mathcal{V_T}$) and learn to synthesize unseen visual features in $\mathcal{V_D}$ for regularization.  
We present the details in the following sections.

\subsection{View Feature Extraction}
\label{ssec:feature_extraction}

We follow the popular multi-view paradigm in 3D object retrieval~\cite{su2015multi}. 
For each 3D object in $\mathcal{O}$ , we first project it into a sequence of $M$ view images $[\mathbf{I}_1,\mathbf{I}_2,...,\mathbf{I}_M] = o$, from different viewpoints folowing~\cite{feng2023hypergraph}, where each view image $\mathbf{I}_m \in \mathbb{R}^{H \times W}$. Next, we feed multi-view images into a pretrained  DINO~\cite{caron2021dinov1,oquab2024dinov2,simeoni2025dinov3} and take out the $d$-dimensional $\mathtt{[CLS]}$ tokens as the global view features. The feature extraction for the multi-view set is defined as $[\mathbf{f}_1, \mathbf{f}_2, \ldots, \mathbf{f}_M] = \mathcal{F}_{\frozen}(o)$, where for each individual view:
\begin{equation}
\mathbf{f}_m = \mathcal{F}_{\frozen}(\mathbf{I}_m), \mathbf{f}_m\in \mathcal{V_D}.
\end{equation}
While any modern pretrained 2D model, especially vision foundation ones (\emph{e.g.}, CLIP~\cite{radford2021learning}), can be used as the vision encoder $\mathcal{F}_{\frozen}(\cdot)$ for high-quality, transferable view features, here we diverge from recent CLIP-based works~\cite{wang2025teda,he2025clip,wang2025describe} based on two considerations: 1) DINO is pretrained with self-supervised learning and provides robust category-agnostic features capturing
global and local cues, which are crucial for generalization to new classes. 2) Our preliminary experiments show that it gives stronger empirical \emph{zero-shot} 3DOR performance than CLIP counterparts.

\subsection{Chunking and Adapting Module}
While mean-pooling across view features provides a decent zero-shot baseline, we consider enhancing it by adapting to a limited set of seen-class 3D objects, mitigating the distribution gap. 
To this end, we design a simple but effective Chunking and Adapting Module (\textbf{CAM}) for adaptation (Figure~\ref{fig:adaptor}), which gradually integrates rich local view relations into compact and discriminative 3D embeddings. Specifically, the whole process is divided into three stages:

\begin{figure}[ht]
\centering
%% old: pipeline_v2
\includegraphics[width=0.95\linewidth]{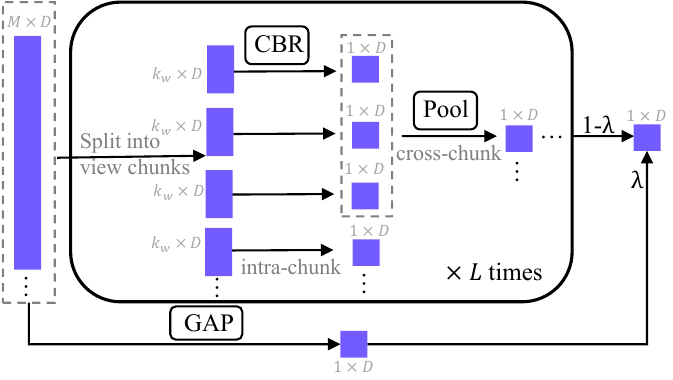}
\caption{\textbf{Illustration of our chunking and adapting module}.}
\vspace{-8pt}
\label{fig:adaptor}
\end{figure}

\noindent\textbf{Local chunk aggregation}. We first divide the view feature sequence $[\mathbf{f}_1,\mathbf{f}_2,...,\mathbf{f}_M]$ into $K$ non-overlapping chunks.
Each chunk comprises $k_w = \lceil M/K \rceil$ consecutive views, with padding applied to the final chunk if $M$ is not divisible by $K$.
Next, we apply a shared linear layer to aggregate each chunk in parallel. The above process could be easily instantiated with a 1D convolution by setting kernel and stride sizes to $k_w$. In particular, we use CBR block (denoted by $\mathtt{CBR}(\cdot)$) to process the whole sequence, with each one consisting of 1D convolutional layer, followed by a batch normalization and a ReLU layer:
\begin{equation}
[\mathbf{g}_1,\mathbf{g}_2,...,\mathbf{g}_K] = \mathtt{CBR}([\mathbf{f}_1,\mathbf{f}_2,...,\mathbf{f}_M]),
\end{equation}
where each $\mathbf{g}^i \in \mathbb{R}^d$ is the $i$-th chunk embeddings derived from $i$-th chunk. 
In this way, we capture local correlations and compositional patterns within a chunk's context. 

\noindent\textbf{Cross-chunk integration}. To integrate cross-chunk features, we apply a non-parametric 1D pooling operation over the chunk embeddings with a predefined pooling window/kernel size, which efficiently captures longer-range dependencies among views:
\begin{equation}
[\mathbf{g}_1', \ldots, \mathbf{g}_{K'}'] = \mathtt{Pool}([\mathbf{g}_1, \mathbf{g}_2, \ldots, \mathbf{g}_K]),
\end{equation}
where $\mathbf{g}_{k'}' \in \mathbb{R}^d$ is the $k'$-th pooled feature and $K'$ is the feature number. By stacking multiple $\mathtt{CBR}(\cdot)$ and $\mathtt{Pool}(\cdot)$ blocks, we progressively extend spatial contexts, ultimately yielding one compact global representation $\mathbf{g}_{\text{adpt}} \in \mathbb{R}^d$. 

\noindent\textbf{Weighted residual fusion}. 
To better preserve the pretrained knowledge and the newly adapted one, we combine them through a weighted residual connection:
\begin{equation}
\mathbf{g}_{\text{final}} = \lambda \cdot \mathbf{f}_{\text{gap}} + (1 - \lambda) \cdot \mathbf{g}_{\text{adpt}},
\end{equation}
where $\mathbf{f}_{\text{gap}} = \mathtt{GAP}([\mathbf{f}_1, \mathbf{f}_2, ..., \mathbf{f}_M])$ denotes the global pretrained feature by global average pooling, $\lambda \in [0, 1]$ is a hyperparamter that balances conserved pre-trained knowledge and the one learned by CAM, for enhancement.

\subsection{Virtual Feature Synthesis}~\label{sec:VFS}
\begin{figure}[!htp]
\centering
%% old: pipeline_v2
\includegraphics[width=0.95\linewidth]{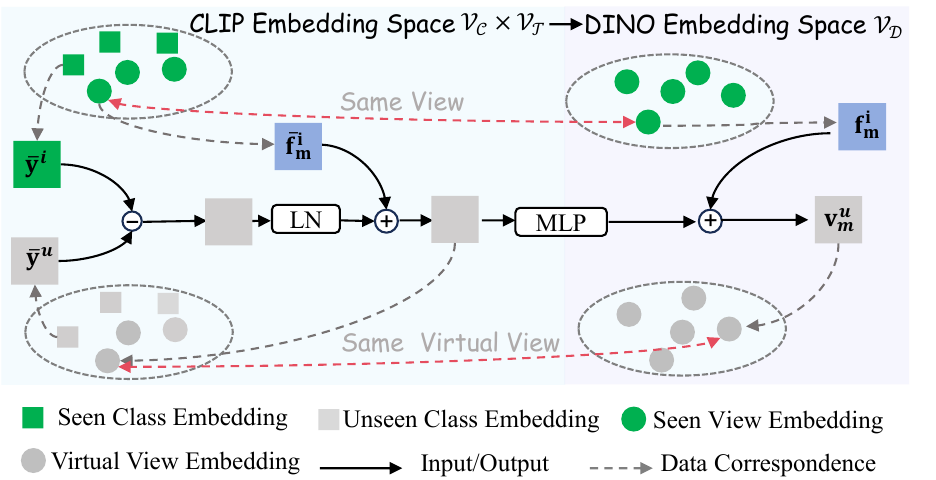}
\caption{\textbf{Illustration of our virtual feature synthesis module}.}
\vspace{-8pt}
\label{fig:unkown_generator}
% \vspace{-7pt}
\end{figure}

To prevent overfitting to ``seen" classes, we design a module named Virtual Feature Synthesis (VFS) to regularize training (Figure~\ref{fig:unkown_generator}). Our key insight is that CLIP's aligned embedding spaces and the semantic relationships between text correlate in the visual space. 
Inspire by it, we propose to synthesize a continuum of virtual visual features for ``unseen" categories by transferring semantic relationships from text modality, thereby encouraging open-set discrimination.

\noindent\textbf{Label space enrichment}.
Our VFS is guided by an extended label space, constructed from a predefined lexicon (\emph{e.g.}, $1k$ categories from ImageNet). 
We filter out overlapping labels with $\mathcal{Y}_{\text{seen}}$.
Meanwhile, to ensure training efficiency, we sub-sample a tractable set of $E$ concepts $\mathcal{Y}_{\text{new}}$. 
The final, enriched concept space is defined as the union of the original and new labels: $\mathcal{Y} = \mathcal{Y}_{\text{seen}} \cup \mathcal{Y}_{\text{new}}$, which provides a scalable foundation for incorporating a much broader universe of semantic concepts.

\noindent\textbf{Aligned embeddings extraction}.
For each view image $\mathbf{I}_m^i$ of a 3D training object $\mathbf{o}^i$, we utilize the CLIP visual encoder to map into aligned visual space $\mathcal{V_C}$, yielding $\mathbf{\overline{f}}^i_m$. For each category $y$, we use the CLIP text encoder to embed the prompt ``a photo of $y$'' into text embedding $\mathbf{\overline{y}} \in \mathcal{V_T}$.
Let $\mathbf{\overline{y}}^i \in \mathcal{V_T}$ denote the corresponding label embedding for label $y_i \in \mathcal{Y}_{\text{seen}}$, and $\mathbf{\overline{y}}^u \in \mathcal{V_T}$ for $y_u \in \mathcal{Y}_{\text{new}}$. We aim to generate a virtual visual embedding $\overline{\mathbf{v}}^{u}_{m} \in \mathcal{V_D}$ for $y_u$. 

\noindent\textbf{Virtual feature synthesis}. 
With aligned space,  we propose to transfer relations between seen and unseen categories—from the textual to the visual modality in CLIP embedding space. Specifically, a virtual visual embedding in the CLIP visual space $\mathcal{V_C}$ is synthesized by shifting a seen-class feature along the semantic direction towards an unseen class. 
It is then projected into the DINO embedding space ($\mathcal{V_D}$) via a lightweight Multi-Layer Perceptron (MLP) $\psi({}\cdot)$, which results in the virtual embedding $\overline{\mathbf{v}}^{u}_{m}$:
\begin{equation}
\overline{\mathbf{v}}^{\text{u}}_m 
= 
\psi[\mathbf{\overline{f}}_m^i 
+\epsilon \cdot \mathtt{LN}(\mathbf{\overline{y}}^u - \mathbf{\overline{y}}^i)],
\label{eq:feature_transformation}
\end{equation}
where  $\epsilon$ denotes a learnable scaling weight, $\mathtt{LN}(\cdot)$ denotes Layer Normalization, and $\mathbf{\overline{y}}^u - \mathbf{\overline{y}}^i$ captures the semantic direction towards the unseen category in $\mathcal{Y}_{\text{new}}$.
This operation assumes that semantic relations are approximately preserved between $\mathcal{V_C}$ and $\mathcal{V_T}$ (see sec.~\ref{sec:theory}).

To stabilize training, we further use a weighted residual fusion, producing final output of VFS is a virtual unseen-class feature $\mathbf{{v}}^{u}_{m} \in \mathcal{V_D}$:
\begin{equation}
\mathbf{{v}}^{u}_{m}=\alpha \cdot\overline{\mathbf{v}}^{u}_{m} + \beta \cdot\mathbf{f}^{i}_{m}.
\label{eq:feature_fusion}
\end{equation}
Here, $\alpha$ and $\beta$ are learnable parameters that balance the contribution between the unseen synthesized features $\overline{\mathbf{v}}^{\text{u}}$ and the original DINO features $\mathbf{f}^{i}$ of the seen-class view images. 

These virtual features of unseen classes serve as regularizers. 
By compelling CAM to discriminate virtual features of unseen classes, we circumvent over-reliance on known class patterns and bolster generalization.
Notably, VFS is only used in training, incurring no extra inference cost.

\subsection{Theoretical Analysis of the VFS module} \label{sec:theory}

Our VFS module employs a common residual connection to merge real features with synthesized ones ( Eq.~\eqref{eq:feature_fusion}), producing virtual features. Its core novelty lies in Eq.~\eqref{eq:feature_transformation}, which utilizes text-visual alignment from CLIP and approximates the expectation of unseen features, $\mathbb{E}[\overline{\mathbf{f}_{m}}|y^u]$, through generalized sampling. Below we give detailed analysis.

\noindent\textbf{CLIP's semantic alignment}.
Being contrastively trained on web-scale text-image pairs, CLIP visual embedding space $\mathcal{V_C}$ and textual embedding space $\mathcal{V_T}$ are well-aligned for broad concepts and visual counterparts.
For a seen category instance, we assume the visual embedding is $\mathbf{\overline{f}}_m^i \approx \overline{\mathbf{y}}^i + \boldsymbol{\eta}$, where $\boldsymbol{\eta}$ represents bounded intra-class noise, which is modeled as a Gaussian distribution $\boldsymbol{\eta} \sim \mathcal{N}(\mathbf{0}, \sigma^2\mathbf{I})$. For an unseen category, assuming access to hypothetical visual embeddings $ \overline{\mathbf{f}}^u_m \approx \overline{\mathbf{y}}^u + \boldsymbol{\eta}'$ (with similar noise $\boldsymbol{\eta}'$), the visual difference is defined as $ \mathbf{\overline{f}}_m^u - \mathbf{\overline{f}
}_m^i$.
Under CLIP's contrastive alignment, the dot product of semantic and visual differences between seen and unseen classes: 
\begin{equation}
\begin{split}
(\overline{\mathbf{y}}^u - \overline{\mathbf{y}}^i)^{\mathtt{T}} (\overline{\mathbf{f}}_m^u - \overline{\mathbf{f}}_m^i) 
= & [(\overline{\mathbf{y}}^u)^{\mathtt{T}}\overline{\mathbf{f}}_m^u + (\overline{\mathbf{y}}^i)^{\mathtt{T}}\overline{\mathbf{f}}_m^i]_{\text{pos}}
\\
& -[(\overline{\mathbf{y}}^u)^{\mathtt{T}}\overline{\mathbf{f}}_m^i + (\overline{\mathbf{y}}^i)^{\mathtt{T}}\overline{\mathbf{f}}_m^u]_{\text{neg}},
\end{split}
\end{equation}
where $[\cdot]_{\text{pos/neg}}$
wraps calculations from positive/negative pairs, respectively. It can be easily observed that semantic relations in $\mathcal{V_T}$ are preserved isometrically in $\mathcal{V_C}$.This provides a principled bedrock for generalized sampling of virtual unseen class features to regularize the model.

\noindent\textbf{Generalized sampling}.
Assuming there exists a Lipschitz mapping $\zeta: \mathcal{V_C} \times \mathcal{V_T} \to \mathcal{V_D}$, with Lipschitz constant $L$, to transform from the CLIP embedding space $\mathcal{V_C} \times \mathcal{V_T}$ to the DINO embedding space $\mathcal{V_D}$. % , preserving the relative distances between the two semantic spaces. 
We regard $\overline{\mathbf{f}}_{m}$ as a random variable in $\mathcal{V_C}$, and $\mathbb{E}[\overline{\mathbf{f}}_{m}|y^u]$ denote the expectation of $\overline{\mathbf{f}}_{m}$ for the u-th unseen class. To facilitate generalized sampling, we employ : $\mathbf{\overline{f}}_m^i 
+\epsilon \cdot \mathtt{LN}(\mathbf{\overline{y}}^u - \mathbf{\overline{y}}^i) \thicksim \mathbb{E}[\overline{\mathbf{f}}_{m}|y^u]$. The learnable parameter $\epsilon$ endows the model with a certain degree of generalization capability. Theoretically, the mapping $\zeta$ exists. In practice, we instantiate it using an MLP ($\psi({}\cdot)$). Through optimization, $\psi({}\cdot)$ is driven towards a reliable approximation. Based on Equation \eqref{eq:feature_transformation} and the Lipschitz property, we have the following proof:
\begin{multline}
\| \overline{\mathbf{v}}^{\text{u}}_m  - \psi(\mathbb{E}[\overline{\mathbf{f}}_{m}|y^u]) \| \\
= \| \psi[\mathbf{\overline{f}}_m^i +\epsilon \cdot \mathtt{LN}(\mathbf{\overline{y}}^u - \mathbf{\overline{y}}^i)] - \psi(\mathbb{E}[\overline{\mathbf{f}}_{m}|y^u]) \| \leq L\cdot \sigma^2,
\end{multline}
where $L$ is the Lipschitz constant, $\sigma^2$ denotes the variance of the noise $\boldsymbol{\eta}$. The result indicates $\overline{\mathbf{v}}^{\text{u}}_m$ is bounded close to $\psi(\mathbb{E}[\overline{\mathbf{f}}_{m}|y^u])$. Such a property enables a trainable $\overline{\mathbf{v}}^{\text{u}}_m$ to regularize for better generalization to unknown classes.

\subsection{Training and Inference}
\label{ssec:training^inference}

To train, we use multi-similarity loss~\cite{wang2019multi}, 
pulling positives together and pushing negatives apart for compact and well-separated classes for both seen and unseen ones.During inference, we use CAM outputs as 3D descriptors for 3DOR.
\section{Experiments}
\label{sec:exp}

\begin{table*}[h]
\centering
\small 
\setlength{\tabcolsep}{1.0pt}
\resizebox{1.0\linewidth}{!}{
\begin{tabular}{l@{\hspace{1pt}}cccc@{\hspace{10pt}}ccc@{\hspace{10pt}}ccc@{\hspace{10pt}}ccc}
\toprule[1pt]
\multirow{2}[2]{*}{Method} & \multirow{2}[1]{*}{Modality} & \multicolumn{3}{c}{OS-ESB-core} & \multicolumn{3}{c}{OS-NTU-core} & \multicolumn{3}{c}{OS-MN40-core} & \multicolumn{3}{c}{OS-ABO-core} \\
\cmidrule(lr){3-5} \cmidrule(lr){6-8} \cmidrule(lr){9-11} \cmidrule(lr){12-14}
 & & \footnotesize{mAP$\uparrow$} & \footnotesize{NDCG$\uparrow$} & \footnotesize{ANMRR$\downarrow$} & \footnotesize{mAP$\uparrow$} & \footnotesize{NDCG$\uparrow$} & \footnotesize{ANMRR$\downarrow$} & \footnotesize{mAP$\uparrow$} & \footnotesize{NDCG$\uparrow$} & \footnotesize{ANMRR$\downarrow$} & \footnotesize{mAP$\uparrow$} & \footnotesize{NDCG$\uparrow$} & \footnotesize{ANMRR$\downarrow$} \\
\midrule
\textbf{~~{Zero-shot baseline}} \\ 
OpenShape (\footnotesize{PB-CLIP B/32})~\cite{liu2024openshape} & P. & 37.64 & 18.38 & 65.57 & 25.53 & 15.41 & 74.51 & 30.31 & 46.34 & 67.01 & 40.29 & 46.09 & 59.42 \\
OpenShape (\footnotesize{PB-CLIP L/14})~\cite{liu2024openshape} & P. & 38.58 & 18.81 & 65.10 & 24.71 & 15.02 & 75.18 & 29.64 & 44.79 & 67.64 & 38.65 & 45.56 & 60.90 \\
ULIP-2 (\footnotesize{PB-CLIP G/14})~\cite{xue2024ulip} & P. & 45.14 & 21.00 & 59.15 & 31.50 & 17.89 & 68.99 & 32.76 & 48.92 & 65.22 & 44.26 & 49.04 & 55.64 \\
Uni3D (\footnotesize{Uni3D-Giant})~\cite{zhou2023uni3d} & P. & 44.42 & 20.92 & 59.96 & 32.02 & 18.04 & 68.49 & 33.21 & 50.51 & 65.11 & 45.92 & 49.79 & 53.96 \\

 CLIP-AdaM (\footnotesize{ViT-B/32})~\cite{he2025clip}  & I., T.  &  53.93 &  23.00 &  49.66 &  49.35 &  23.70 &   53.03  
 & 49.60 &   65.73 &  50.89 
 &  47.73&   51.76 &  52.48 \\
 CLIP-AdaM (\footnotesize{ViT-L/14})~\cite{he2025clip} & I., T.  & 54.69 & 23.40 & 48.68 & 57.28 & 26.24 & 45.47 & 55.01 &  70.56 & 45.72 & 57.29 &   56.13 & 44.51 \\

\cc Our baseline (DINOv2 ViT-B/14)  & \cc I. & \cc \underline{59.19} & \cc 24.13 & \cc \underline{44.31} & \cc 59.77 & \cc 26.77 & \cc 43.25 & \cc 62.77 & \cc 74.76 & \cc 39.42 & \cc 62.17 & \cc 57.93 & \cc 40.23 \\
\cc Our baseline (DINOv2  ViT-L/14)  & \cc I. & \cc 58.25 & \cc \underline{24.24} & \cc 46.00 & \cc \underline{63.24} & \cc \underline{27.73} & \cc \underline{39.73} & \cc \underline{65.42} & \cc \underline{77.06} & \cc \underline{36.72} & \cc \underline{64.92} & \cc \underline{58.85} & \cc \underline{37.70} \\
\cc Our baseline (DINOv3 ViT-B/16)  & \cc I. & \cc 57.93 & \cc 23.86 & \cc 45.85 & \cc 59.33 & \cc 26.49 & \cc 43.95 & \cc 62.25 & \cc 75.71 & \cc 39.76 & \cc 61.84 & \cc 57.55 & \cc 40.39 \\
\cc  Our baseline (DINOv3 ViT-L/16)  & \cc I. & \cc \textbf{61.44} & \cc \textbf{24.88} & \cc \textbf{42.63 }& \cc \textbf{66.34} & \cc \textbf{28.45} & \cc \textbf{36.79 }& \cc \textbf{68.18} & \cc \textbf{79.25 }& \cc \textbf{34.31} & \cc \textbf{67.02} & \cc \textbf{60.00} & \cc \textbf{35.34} \\

\midrule 
\textbf{{~~Open-set 3DOR}} \\ 
TCL~\cite{he2018triplet} & P., I., V.  
& 49.31 & 21.89 & 52.68 & 39.37 & 21.23 & 61.00 & 48.11 & 63.83 & 52.30 & 49.33 & 53.86 & 51.05 \\
SDML~\cite{hu2019scalable} & P., I., V.  
& 49.59 & 21.75 & 52.36 & 40.16 & 21.52 & 60.49 & 50.75 & 65.70 & 50.22 & 47.44 & 52.79 & 52.42 \\
CMCL~\cite{jing2021cross} & P., I., V.   
& 50.01 & 21.97 & 53.06 & 41.08 & 21.72 & 59.43 & 51.38 & 65.98 & 49.75 & 49.83 & 50.89 & 50.24 \\
MMSAE~\cite{wu2019multi} & P., I., V. 
& 49.88 & 22.06 & 53.69 & 40.85 & 21.70 & 59.99 & 52.08 & 66.57 & 49.00 & 50.51 & 53.80 & 50.49 \\
MCWSA~\cite{zheng2022multi} & P., I., V. 
& 49.48 & 21.34 & 53.75 & 39.22 & 20.69 & 62.14 & 48.78 & 63.85 & 51.95 & 45.61 & 51.05 & 54.70 \\
PROSER~\cite{zhou2021learning} & P., I., V. 
& 48.69 & 21.13 & 53.95 & 39.47 & 21.24 & 60.96 & 49.00 & 64.54 & 51.66 & 50.33 & 53.27 & 50.34 \\
InfoNCE~\cite{oord2018representation}  & P., I., V. 
& 50.26 & 21.91 & 52.63 & 40.03 & 21.19 & 61.09 & 47.37 & 63.31 & 53.02 & 46.83 & 52.14 & 53.50 \\
HGM$^{2}$R~\cite{feng2023hypergraph} & P., I., V. 
& 51.74 & 22.73 & 51.28 & 44.88 & 22.81 & 56.67 & 64.20 & 72.91 & 38.27 & 63.39 & 57.96 & 37.96 \\
CLIP-AdaM(\footnotesize{ViT-B/32})~\cite{he2025clip} & I., T. & 57.23 & 24.05 & 47.73 & 52.47 & 24.77 & 50.97 & 57.09 & 70.13 & 44.70 & 57.65 & 57.52 & 43.37 \\
CLIP-AdaM(\footnotesize{ViT-L/14})~\cite{he2025clip} & I., T. & 59.90 & \underline{24.69} & 44.80 & 61.52 & 27.65 & 41.72 & 63.54 & 74.21 & 38.76 & 64.77 & 58.75 & 37.25 \\
DAC(\footnotesize{CLIP ViT-B/32})~\cite{wang2025describe} & I., T. & 58.70 & 24.27 & 45.67 & 59.21 & 27.06 & 44.58 & 62.40 & 72.63 & 39.82 & 66.10 & 59.01 & 36.12 \\
DAC(\footnotesize{CLIP ViT-L/14})~\cite{wang2025describe}  &  I., T. & 57.80 &24.36 &47.44 & \underline{65.83} & \underline{28.78} & \underline{37.46} &68.98 & 77.59 &33.87 & \underline{70.74} & \underline{60.87} & \underline{32.14} \\
\cc Ours(DINOv2 ViT-B/14) & \cc I. & \cc \underline{61.82} & \cc 24.55 & \cc \underline{42.74} & \cc 61.56 & \cc 27.27 & \cc 41.62 & \cc 67.62 & \cc 77.67 & \cc 34.95 & 
%\cc 65.01 & \cc 59.31 & \cc 37.35 
\cc 65.04 & \cc 59.34 & \cc 37.32\\
\cc Ours(DINOv2 ViT-L/14)  & \cc I. & \cc 60.48 & \cc 24.47 & \cc 43.25 & \cc 65.15 & \cc 28.24 & \cc 37.80 & \cc %69.17
\underline{69.69} & \cc %78.68
\underline{79.11} & \cc %33.73
\underline{33.23} & \cc %67.10
67.91& \cc  %60.67
60.73& \cc %35.94
34.74\\
\cc Ours(DINOv3 ViT-B/16)  & \cc I. & \cc 59.06 & \cc 24.08 & \cc 44.77 & \cc 60.01 & \cc 26.91 & \cc 43.21 & \cc 65.44 & \cc 77.75 & \cc 36.96 & \cc 63.66 & \cc 58.74 & \cc 38.94 \\
\cc Ours(DINOv3 ViT-L/16)  & \cc I. & \cc  \textbf{62.75} & \cc \textbf{25.29} & \cc  \textbf{42.33} & \cc \textbf{67.75} & \cc \textbf{28.92} & \cc \textbf{35.26} & \cc \textbf{72.15} & \cc \textbf{81.18} & \cc \textbf{30.71} & \cc \textbf{70.96} & \cc \textbf{61.69} & \cc \textbf{31.53} \\

\bottomrule[1pt]
\end{tabular}
}

\caption{\textbf{Performance comparisons (\%) on open-set 3DOR benchmarks}. 
\textbf{Bold} and \underline{underline} indicate the best and second best results, respectively. 
PB-CLIP refers to PointBERT-CLIP. P., I., V., and T. stand for Point Cloud, Multi-view Images,Voxel and Text, respectively. 
}
\label{tab:main_results}
\vspace{-12pt}
\end{table*}

\subsection{Experimental Setup}

\noindent\textbf{Datasets and evaluation metrics}.
We evaluate on four standard open-set 3DOR datasets~\cite{feng2023hypergraph}, including OS-ESB-core, OS-NTU-core, OS-MN40-core, and OS-ABO-core.
Following common practice~\cite{feng2023hypergraph, wang2025describe}, we use mean Average Precision (mAP), Normalized Discounted Cumulative Gain (NDCG) and Average Normalized Modified Retrieval Rank (ANMRR) as evaluation metrics.

\noindent\textbf{Implementation details}.
For a fair comparison, we adopt the same projection scheme as HGM2R~\cite{feng2023hypergraph, he2025clip, wang2025describe}, obtaining 24-view images at $256\times256$ resolution for each 3D object.
The CBR block  $\mathtt{CBR}(\cdot)$ adopts a 3 kernel with stride 3, and $\mathtt{Pool1D}(\cdot)$ a kernel 3.
We extensively benchmark with four pre-trained DINO models, DINOv2 ViT-B/14, DINOv2 ViT-L/14, %DINO.TXT ViT-L/14, 
DINOv3 ViT-B/16, and DINOv3 ViT-L/16. 
In addition, the Virtual Feature Synthesis (VFS) module uses the pretrained CLIP ViT-B/16 model.
We use SGD with momentum 0.9 and weight decay $5 \times 10^{-4}$ for optimization.
The adapter is trained with a learning rate of $1{\times}10^{-3}$, while VFS is initialized at $1{\times}10^{-4}$.
Each mini-batch contains 8 samples from 2 categories, with 4 instances per category.
We train for 70 epochs and decay the learning rate by a factor of 0.1 at the 20th and 40th epochs.
We use the ImageNet label set as our lexicon for unseen classes.
All experiments are conducted on a single RTX 4090 GPU. 
Please refer to \underline{\textit{Appendix}} for detailed model settings.

\subsection{Comparison with state-of-the-arts}
\noindent\textbf{Comparison methods.}
We compare DEC extensively against state-of-the-art methods, which we categorize as follows: (1) Traditional 3DOR methods, including TCL~\cite{he2018triplet}, SDML~\cite{hu2019scalable}, CMCL~\cite{jing2021cross}, MMSAE~\cite{wu2019multi}, MCWSA~\cite{zheng2022multi}, InfoNCE~\cite{oord2018representation}. These works utilize multi-modalities for generalized 3D descriptors. 
(2) Large-scale zero-shot models, including some recent models pre-trained on large-scale point cloud datasets, such as OpenShape~\cite{liu2024openshape}, ULIP-2~\cite{xue2024ulip}, and Uni3D~\cite{zhou2023uni3d}.
(3) Hypergraph-based method HGM$^2$R that model train-test data relations using multi-modalities.
(4) Recent CLIP-based methods, CLIP-AdaM~\cite{he2025clip} and DAC~\cite{wang2025describe}.
CLIP-AdaM carefully designs an adapter after CLIP, and DAC synergizes MLLM-inferred text with multi-view images in a CLIP-based framework.
In contrast, our DEC operates solely on multi-view images, making it more scalable and efficient.

\noindent\textbf{Result analysis.}
Table~\ref{tab:main_results} summarizes results under both zero-shot and open-set settings. 
Under the zero-shot setup, we establish a simple multi-view DINO baseline by mean-pooling the multi-view features for retrieval. 
Notably, this simple baseline substantially outperforms all the compared methods, covering CLIP-based and point-cloud-based methods, across multiple pretrained DINO variants.   
For instance, with the DINOv3 ViT-L/14 backbone, it achieves over 10\% mAP improvement on OS-MN40-core %and nearly 8\% mAP on OS-ABO-core, 
surpassing the best CLIP-based baseline by a large margin.
Notably, on OS-ESB-core, which comprises high-genus objects like mechanical parts~\cite{jayanti2006developing} and requires discerning subtle structural nuances, our established baseline surpasses all specifically designed open-set methods, even those that require training (\emph{e.g.}, DAC~\cite{wang2025describe}). 
It demonstrates that DINO’s inherent ability to capture fine-grained local details yields more robust and generalizable features, thereby benefiting further adaptation in open-set 3DOR.

In the open-set setting, with only a limited number of 3D objects from seen categories available, DEC further enhances generalization through adaptation, advancing state-of-the-art greatly across all benchmarks.
For example, compared with DAC~\cite{wang2025describe}, the current SOTA in open-set 3DOR, although we \emph{only} utilize multi-view images, our method (DINOv2 ViT-B/14) improves mAP over DAC (CLIP ViT-B/32) by +3.12\% (on OS-ESB-core), +2.35\% (on OS-NTU-core), and +5.22\% (on OS-MN40-core), three of the four standard benchmarks, respectively. 
More importantly, when equipped with stronger backbones, further performance gain can be acquired for our method.  
For example, compared with DAC (CLIP ViT-L/14), our method (DINOv3 ViT-L/16) improves mAP by +0.22\%, NDCG by up to +0.88\%, and reduces ANMRR by -0.61\% on OS-ABO-core.
The consistent improvements provide strong evidence for the efficacy of DEC across a spectrum of benchmarks—from fine-grained mechanical parts (OS-ESB-core) to large-scale common objects (OS-MN40-core and OS-NTU-core), and real-world products (OS-ABO-core).

\subsection{Component Analysis}
All ablations are conducted on OS-MN40-core.
For more analysis, please refer to \underline{\textit{Appendix}}.

\noindent\textbf{Effectiveness of components.}
We conduct a step-wise ablation with two DINOv2 variants on the core components (CAM and VFS) of our framework. 
As shown in Table~\ref{tab:ablation}, our \emph{baseline} that uses a three-layer MLP adapter over the mean-pooled DINO view features, yields limited performance. Specifically, with the ViT-B/14 backbone, it only reaches 63.49\%, 75.02\%, 38.86\% in mAP, NDCG, ANMRR, respectively. Replacing it with CAM greatly increases mAP and NDCG by +2.94\% and +1.78\%, while reducing ANMRR by -0.91\%. 
This indicates that by integrating local view relations gradually, CAM effectively produces more discriminative and generalized features. 
When combined with VFS, further notable gains are observed. For example, with DINOv2 ViT-B/14, mAP goes from 66.43\% to 67.62\%, demonstrating the complementary enhancement of virtual features in open-set generalization.

\begin{table}[t]
\centering
\tablestyle{10pt}{1.1}
\small
\setlength{\tabcolsep}{6pt}
\begin{tabular}{lccc ccc}
\toprule[1pt]
Backbone & CAM & VFS & mAP$\uparrow$ & NDCG$\uparrow$ & ANMRR$\downarrow$ \\
\midrule
\multirow{3}{*}{ViT-B/14}
&\xmark & \xmark &  63.49 & 75.02 & 38.86\\
 & \cmark & \xmark & 66.43 & 76.80 & 37.95 \\
 & \cmark & \cmark & \textbf{67.62} & \textbf{77.67} & \textbf{34.95} \\
\midrule
\multirow{3}{*}{ViT-L/14}
&\xmark & \xmark &  65.89   &  77.29  &  36.55 \\
 & \cmark & \xmark & 68.43 & 78.50 & 34.56 \\
 & \cmark & \cmark & \textbf{69.69} & \textbf{79.11} & \textbf{33.23} \\
\bottomrule[1pt]
\end{tabular}
\caption{\textbf{Ablations of components on OS-MN40-core.} }

\vspace{-8pt}
\label{tab:ablation}
\end{table}

\begin{figure*}[t]
%\vspace{-0.3in}
\centering
\begin{minipage}{0.66\textwidth}
\centering
\begin{minipage}{0.48\textwidth}
\includegraphics[width=\linewidth]{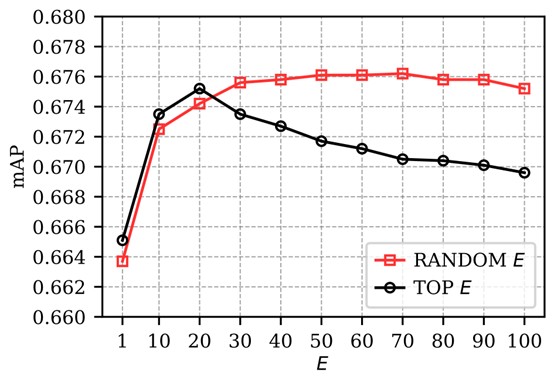} % org: pdf
\end{minipage}
\hfill
\begin{minipage}{0.48\textwidth}
\includegraphics[width=\linewidth]{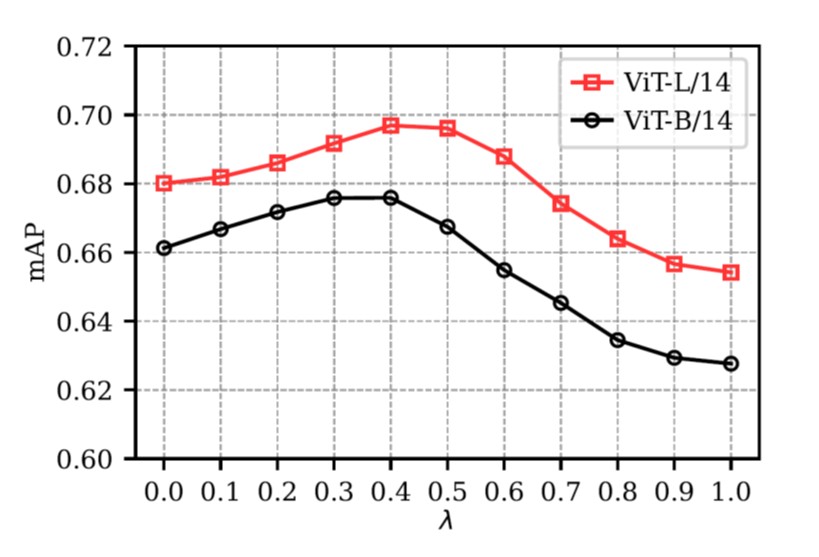} % org: pdf
\end{minipage}
\vspace{-8pt}
\caption{Performance comparison on OS-MN40-core. \emph{Left:} Selection schemes with \\ different $E$ value. \emph{Right:} Impact of fusion weight $\lambda$.}
\label{fig:hyper_k_fusion}
\end{minipage}
\hfill
\begin{minipage}{0.33\textwidth}
\centering
\vspace{0.05cm}
\includegraphics[width=0.96\linewidth]{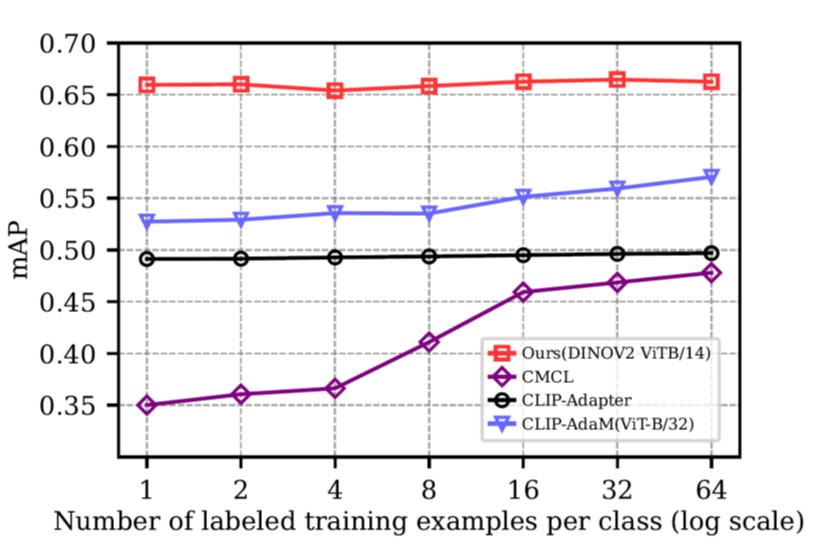} % org: pdf
\vspace{-8pt}
\caption{Impact of training sample numbers per class on OS-MN40-core.}
\label{figure:fewshot}
\end{minipage}
\vspace{-12pt}
\end{figure*}

\noindent\textbf{The influence of $E$ under different selection schemes. }
To further examine the influence of the sub-sampling strategy in our Virtual Feature Synthesis (VFS) module (Sec.~\ref{sec:VFS}), we compare two concept selection schemes: \textit{1) RANDOM $E$}, which uniformly samples $E$ unseen concepts from the category set ${\mathcal{Y}_{\text{new}}}$, introducing stochastic semantic variations; and \textit{2) TOP $E$}, which measures the cosine similarity between the CLIP image feature and all unseen textual embeddings, selecting the top $E$ closest ones to ensure semantic coherence between visual and textual spaces.
The experiments are performed on OS-MN40-core using DINOv2 ViT-B/14 as the backbone. As shown in the left panel of Figure~\ref{fig:hyper_k_fusion}, and show that the \textit{RANDOM~$E$} selection scheme gradually increases with $E$, and then saturating at $40$. \emph{TOP E} performs better than the random scheme for smaller $E$, but it rapidly introduces redundant, semantically similar concepts, which hinders further learning. Conversely, RANDOM $E$ benefits from stochastic semantic diversity to explore broader unseen space, which acts as a regularization and promotes generalization.

\noindent\textbf{Comparative analysis of adapters.}
We further compare our adapter with other counterparts, including CLIP-Adapter~\cite{gao2024clip} and CLIP-AdaM~\cite{he2025clip}. As shown in Table~\ref{tab:adapter_methods}, %the baseline model without any adapter achieves 62.77\%, 74.76\%, and 39.42\% in mAP, NDCG, and ANMRR, respectively. 
compared with CLIP-Adaptor~\cite{gao2024clip} and CLIP-AdaM~\cite{he2025clip}, our CAM achieves additional gains.
This advantage arises from CAM’s localized chunk aggregation, which captures inter-view structural dependencies that are often overlooked by global pooling approaches.
By promoting coherent yet compact representations, CAM produces embeddings that generalize more effectively across unseen categories and exhibit improved stability under open-set setting.

\begin{table}[t]
    \centering
     \tablestyle{10pt}{1.1}
    \begin{tabular}{lccc}\toprule[1pt]
         Method&  mAP$\uparrow$&  NDCG$\uparrow$ & ANMRR$\downarrow$ \\\midrule
         % Baseline&  62.77&  74.76& 39.42\\
 CLIP-Adaptor~\cite{gao2024clip}& 64.35& 75.56&37.95\\
 CLIP-AdaM~\cite{he2025clip}& 64.46& 75.64&38.13\\
         \cc{Ours}&  \cc\textbf{66.43}&  \cc\textbf{76.80}& \cc\textbf{36.25}\\ \bottomrule[1pt]
    \end{tabular}
    \caption{\textbf{Performance of different adapters on OS-MN40-core.}}
    \label{tab:adapter_methods}
    \vspace{-8pt}
\end{table}

\noindent\textbf{Impact of fusion weight $\lambda$}.
We investigate how the fusion weight $\lambda$ affects the balance between frozen DINO representations and adapted features. As shown in the right panel of Figure~\ref{fig:hyper_k_fusion}, both ViT-B/14 and ViT-L/14 backbones exhibit a similar trend. 
When $\lambda$ increases from 0.0 to 0.4, performance improves steadily, with mAP reaching its peak at around $\lambda=0.4$. 
This suggests that a moderate fusion allows the adapter to enhance DINO features while preserving the pretrained priors.
Beyond this point, further increasing $\lambda$ causes a gradual decline in performance, as the model begins to overfit to the adapted space and loses the robust geometric priors encoded in DINO.
Moreover, the larger ViT-L/14 backbone achieves consistently higher mAP and exhibits a smoother performance curve, indicating that a stronger pretrained backbone not only yields more discriminative and robust features but also provides greater tolerance to feature adaptation. In conclusion, these results highlight that $\lambda$ controls the balance between representation stability and adaptation flexibility, and the best performance appears when the fusion allows the adapter to complement, rather than override, the pretrained DINO representations.

\noindent\textbf{Analysis of the CBR block.} We conduct comprehensive experiments to evaluate the impact of kernel size in the CBR block (Sec.~\ref{sec:method}). As shown in Table~\ref{tab:CBR_methods}, the chunk size of 3 achieves the best results across all metrics. %, with \textbf{67.62\%} mAP, \textbf{77.67\%} NDCG, and \textbf{34.95\%} ANMRR. 
We argue that this is due to a more suitable receptive field, and too large chunks also weaken discriminative local patterns. %However, 

\begin{table}[t]
    \centering
    \tablestyle{10pt}{1.1}
    \begin{tabular}{cccc}\toprule[1pt]
         chunk size& mAP$\uparrow$&  NDCG$\uparrow$ & ANMRR$\downarrow$ \\\midrule
     1   &  63.62 &  75.38& 38.77\\
 3 &\textbf{67.62}& \textbf{77.67}& \textbf{34.95}\\
5 & 64.91& 75.68& 37.08\\
7 &  65.72&  76.42&  36.87\\ \bottomrule[1pt]
    \end{tabular}
    \caption{\textbf{Impact of different chunk sizes
    on OS-MN40-core.}}
    \label{tab:CBR_methods}
    \vspace{-8pt}
\end{table}

\subsection{Extending to More Challenging Scenarios}

\noindent\textbf{Cross-dataset open-set 3DOR.}
To evaluate the cross-dataset generalization, we train DEC on OS-MN40-core and evaluate on OS-ABO-core. 
As shown in Table~\ref{tab:cross-dataset}, DEC with ViT-B/14 already achieves competitive performance, surpassing several multi-modal baselines that utilize point cloud, multi-view images, and voxel modalities. With the stronger ViT-L/16 backbone, DEC achieves an mAP of 69.12\% and an NDCG of 60.44\%, demonstrating strong cross-dataset retrieval capabilities.
Although DAC (ViT-L/14) slightly outperforms DEC in mAP (69.86\%), it leverages Multi-Modal Large Language Models (MLLM) to generate textual embeddings, providing additional semantic guidance for unseen categories. In contrast, DEC relies solely on visual features, yet still maintains comparable performance, highlighting the robustness and generalizability of the representations learned by our framework.

\begin{table}[ht]
\centering
\small
\setlength{\tabcolsep}{14pt}
\resizebox{1.0\linewidth}{!}{
\begin{tabular}{lccccc}
\toprule[1pt]
\multirow{2}{*}{Method} &  \multirow{2}{*}{Modality} 
& \multicolumn{3}{c}{OS-MN40-core $\rightarrow$ OS-ABO-core} \\
 & & mAP$\uparrow$ & NDCG$\uparrow$ & ANMRR$\downarrow$  \\
\midrule
CMCL~\cite{jing2021cross}   & P., I., V.  & 53.90 & 53.79 & 47.28  \\
MMSAE~\cite{wu2019multi}  & P., I., V.  & 52.83 & 52.80 & 48.02  \\
MCWSA~\cite{zheng2022multi}  &P., I., V.  &  49.20 & 50.99 & 51.11 \\
PROSER~\cite{zhou2021learning} &  P., I., V. & 50.80 & 52.37 & 49.73  \\
InfoNCE~\cite{oord2018representation}& P., I., V. &  51.63 & 52.75 & 49.16 \\
HGM$^{2}$R~\cite{feng2023hypergraph}&  P., I., V. & 57.55 & 54.14 & 45.35 \\
DAC (ViT-B/32)&  I., T. & 63.45 & 57.34 & 38.48   \\
DAC (ViT-L/14) & I., T.  & \textbf{69.86} & \underline{60.13} & \textbf{32.42}   \\
\cc Ours (ViT-B/14)& \cc I. & \cc 63.49 & \cc 57.39 & \cc 38.73   \\
\cc Ours (ViT-L/16) &  \cc I.   & \cc \underline{69.12} & \cc \textbf{60.44} & \cc \underline{32.50}   \\
\bottomrule[1pt]
\end{tabular}
}
\caption{\textbf{Comparisons (\%) on cross-dataset retrieval}.}
\label{tab:cross-dataset}
\vspace{-8pt} 
\end{table}

\noindent{\textbf{Few-shot open-set 3DOR.}}
To further evaluate the robustness of DEC under limited supervision, we conduct few-shot experiments on OS-MN40-core, varying the number of labeled samples per class from 1 to 64. As shown in Figure~\ref{figure:fewshot}, DEC consistently outperforms previous arts CMCL~\cite{jing2021cross}, CLIP-Adapter~\cite{gao2024clip}, and CLIP-AdaM~\cite{he2025clip} across all shot settings. The performance gap is most evident in the very low-shot regime (1–4 shots), where all methods perform relatively poorly, but DEC maintains a clearly higher mAP. 
As sample number increases, DEC still performs the best, indicating that it effectively leverages limited data to build a stable and transferable feature space.

\subsection{Visualization and Limitations}
\begin{figure}[ht]
\vspace{-4pt}
\centering
\includegraphics[width=\linewidth]{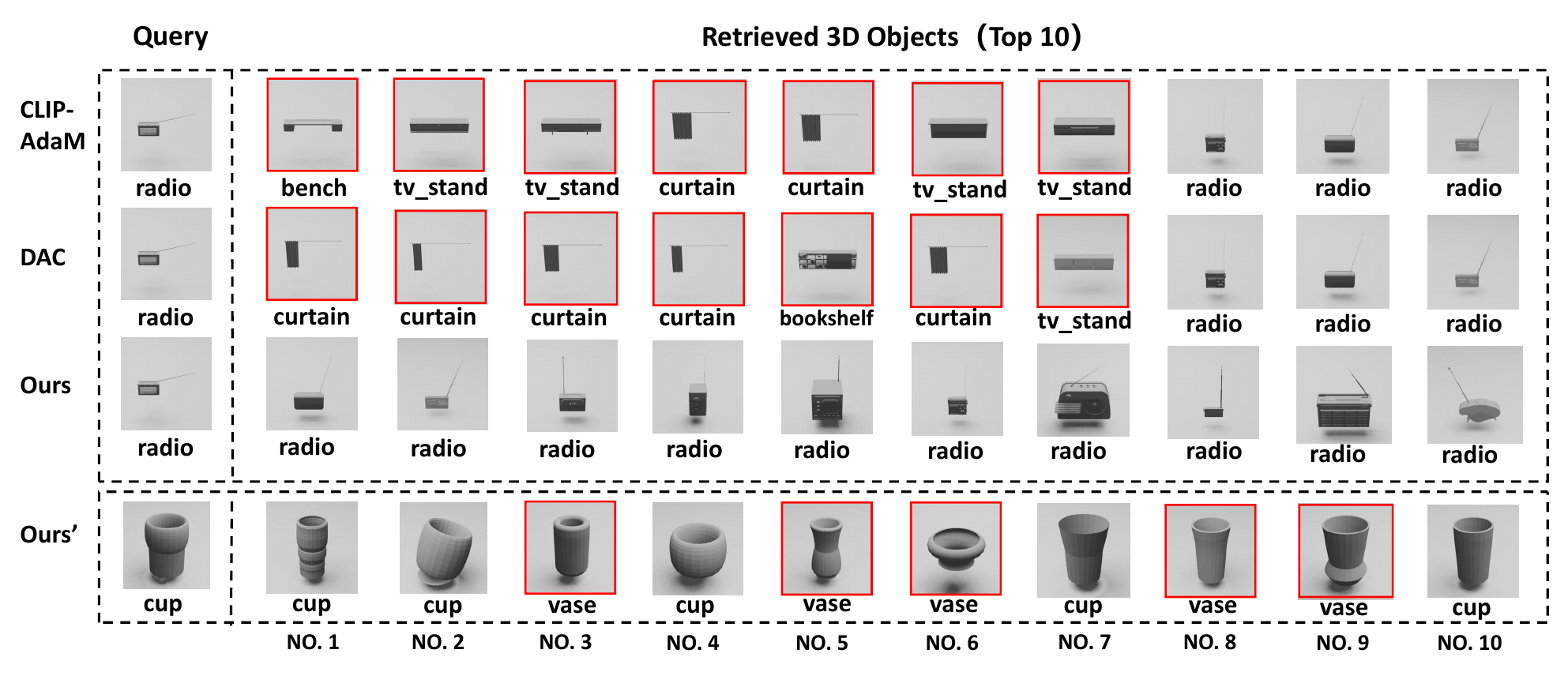}
\caption{\textbf{Retrieval example comparisons with other methods on OS-MN40-core.} Incorrect matches are in red boxes.}
\label{fig:retrieval}
\end{figure}

Despite its SOTA performance, our method has limitations. As shown in Figure~\ref{fig:retrieval} for more qualitative examples, while it captures fine-grained details effectively (e.g., for small objects like a radio), it is sometimes confused by symmetric objects like cup and vase. 
This suggests a need for more careful balancing between global shape and local detail. See \underline{\textit{Appendix}} for more qualitative samples.

\section{Conclusions}
\label{sec:conclusion}

In this paper, we presented a new framework, DINO Eats CLIP (DEC), to adapt multi-view images for open-set 3DOR. 
Building upon the robust self-supervised view representations, we designed a novel chunking and adapting module to effectively integrates rich local relations across views into compact and generalized features. To mitigate overfitting to seen classes, we further introduced a CLIP-based virtual feature synthesizer to regularize the training. Through extensive experiments, we validated its superiority on open-set 3DOR. We hope DEC can inspire research in generalized representation learning beyond 3D domain.

%% funding information
\section*{Acknowledgments}
This work was supported by the National Natural Science Foundation of China (No. 62302188, 62225603).

{
    \small
    \bibliographystyle{ieeenat_fullname}
    \bibliography{main}
}

\clearpage
\setcounter{page}{1}
\maketitlesupplementary

\appendix

In the supplementary material, we first provide more implementation details. Next, we conduct more analysis experiments. Then, we provide more qualitative results. 
Lastly, we evaluate DEC on more challenging and realistic scenarios. 

\section{More Implementation Details}

We set the chunk size to 3, efficiently reducing the number of views by a factor of 6 with each CBR block and pooling layer. 
Thus, by design, CAM is highly light-weight that comprises only maximum of two CBR blocks to process an initial set of 24 views. 
Empirically, a single block is optimal for smaller datasets (OS-ESB-core, OS-NTU-core), while two blocks yield stronger performance on larger ones (OS-MN40-core, OS-ABO-core). When only one block is used, we simply aggregate the view features into a compact representation using adaptive average pooling.

We also implement a three-layer MLP as our plain adapter for our baseline, which follows an encoder-decoder architecture:  the first layer projects the input dimension $D$ down to a bottleneck dimension $D/4$, the second layer operates within this bottleneck, and the third layer projects the features back to the original dimension $D$ ($D=768$).

\section{More Analyses}
\label{sec:more_ablation}
\begin{figure}[ht]
\centering
\includegraphics[width=\linewidth]{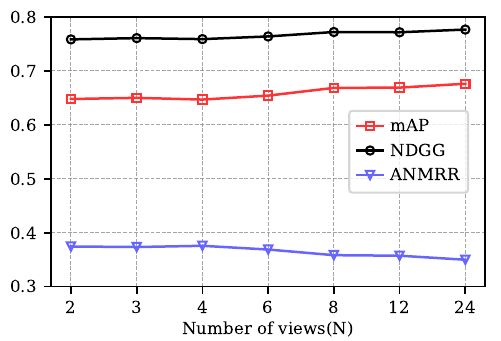}
\caption{\textbf{Effect of view numbers (N) on OS-MN40-core.}}
\label{fig:view_number}
\vspace{-12pt}
\end{figure}

\noindent\textbf{Impact of View Number.}
We analyze the impact of view number on OS-MN40-core. 
As shown in Figure~\ref{fig:view_number}, performance remains relatively stable when using only 2 to 4 views, then increases markedly between 4 and 8 views, and continues to improve slightly up to 24 views. This trend indicates that our method can effectively use a moderate number of views for decent performance. 
Yet we observe diminishing returns once most geometric view information has been observed.

\noindent\textbf{Impact of fusion weight $\lambda$.}
Table~\ref{tab:fusion_weight} summarizes optimal $\lambda$ across benchmarks and backbones.
For OS-ESB-core and OS-NTU-core, the optimal $\lambda$ remains consistently small (0.11) for both ViT-B/14 and ViT-L/14, suggesting that the pretrained DINO embeddings already encode sufficiently discriminative geometric information for these datasets. Consequently, only a minimal contribution from the adapted feature branch is required to attain peak performance, and excessive adaptation may perturb the pretrained feature structure.
Conversely, OS-MN40-core and OS-ABO-core exhibit substantially higher optimal fusion weights, ranging from 0.2 to 0.4. These datasets involve greater intra-class variability and more heterogeneous shape distributions, thereby increasing the necessity for task-specific feature adjustment. In such settings, the adapter contributes more complementary information, and the model benefits from assigning a larger relative weight to the adapted features.

\begin{table}[h]
    \centering
    \resizebox{\linewidth}{!}
    {
        \begin{tabular}{cccccc}
        \toprule
        Dataset & Backbone & $\lambda$& mAP$\uparrow$ & NDCG$\uparrow$ & ANMRR$\downarrow$ \\
        \midrule
        \multirow{2}{*}{OS-ESB-core} 
            & ViT-B/14& $\lambda=0.11$& 61.82& 24.55& 42.74\\
            &             ViT-L/14 & $\lambda=0.11$& 60.59& 24.46& 43.30\\
            \cmidrule{1-6}
        \multirow{2}{*}{OS-NTU-core} 
            & ViT-B/14& $\lambda=0.11$& 61.56& 27.26& 41.62\\
            & ViT-L/14 & $\lambda=0.11$& 65.15& 28.24& 37.80\\
            \cmidrule{1-6}
        \multirow{2}{*}{OS-MN40-core} 
            & ViT-B/14& $\lambda=0.3$& 67.62& 77.67& 34.99\\
            & ViT-L/14 & $\lambda=0.4$& 69.69& 79.11& 33.23\\
            \cmidrule{1-6}
        \multirow{2}{*}{OS-ABO-core} 
            & ViT-B/14& $\lambda=0.3$& 65.04& 59.34& 37.32\\
            & ViT-L/14 & $\lambda=0.2$& 67.91& 60.73& 34.74\\
        \bottomrule
        \end{tabular}
    }
    \caption{\textbf{Optimal $\lambda$ across different datasets and backbones.} }
    \label{tab:fusion_weight}
\end{table}

\noindent\textbf{Overlapping vs. Non-overlapping Chunking.}
Table~\ref{tab:stride_methods} studies the impact of stride size in CBR. The non-overlapping configuration (stride = 3)  yields the highest performance across mAP, NDCG, and ANMRR. It indicates that non-overlapping chunking is not only sufficient but also works better: it successfully captures diverse local contexts while avoiding the computational redundancy and potential feature dilution associated with overlapping strides. Hence, DEC's non-overlapping chunking strategy achieves a much better balance between effectiveness and efficiency.
\begin{table}[t]
    \centering
    \tablestyle{10pt}{1.1}
    \small 
        \resizebox{\linewidth}{!}
    {
    \begin{tabular}{cccc}\toprule[1pt]
         stride size& mAP$\uparrow$&  NDCG$\uparrow$ & ANMRR$\downarrow$ \\
         \midrule
         1   &  64.64 &  76.31 &  37.34\\
         2 & 65.52 &76.62  & 36.74\\
         3 & \textbf{67.62} & \textbf{77.67} & \textbf{34.95}\\
   \bottomrule[1pt]
 \end{tabular}}
    \caption{\textbf{Impact of different stride sizes
    on OS-MN40-core.}}
    \label{tab:stride_methods}
\end{table}

\noindent\textbf{Combining DINO and CLIP Features}
To study whether directly combining CLIP and DINO features derives more effective representations, we train an MLP on added DINOv2-ViT/B14 (mapped to the same dim. as CLIP) and CLIP ViT/B16 features with the standard cross-entropy loss. 
Yet it is worse than the DINOv2 baseline (Table~\ref{tab:baseline}) on 3/4 datasets in mAP. 
It suggests that naive fusion struggles to reconcile the two divergent feature spaces, whereas ours uses CLIP to provide semantic, generalized priors that regularize and guide DINO’s training, resulting in a more effective synergy.

\begin{table}[!ht]
\centering
\tablestyle{10pt}{1.1}
\small
\resizebox{\linewidth}{!}{%
\begin{tabular}{lccccc}
\toprule[1pt]
Methods  &  OS-ESB & OS-NTU & OS-MN40 & OS-ABO \\
\midrule
DINOv2 & 59.19 & 59.77 & 62.77 & 62.17 \\
\textit{Base.} (DINOv2+CLIP) & 59.40 & 59.28 & 61.35  & 60.69  \\
Ours & \textbf{61.82} & \textbf{61.56} & \textbf{67.62} & \textbf{65.04} \\
\bottomrule[1pt]
\end{tabular}}

\caption{\textit{Base.} fuses DINO and CLIP, while we use DINO only.}
\label{tab:baseline}
\end{table}

\section{More Qualitative Results}

\begin{figure}[ht]
\centering
\includegraphics[width=\linewidth]{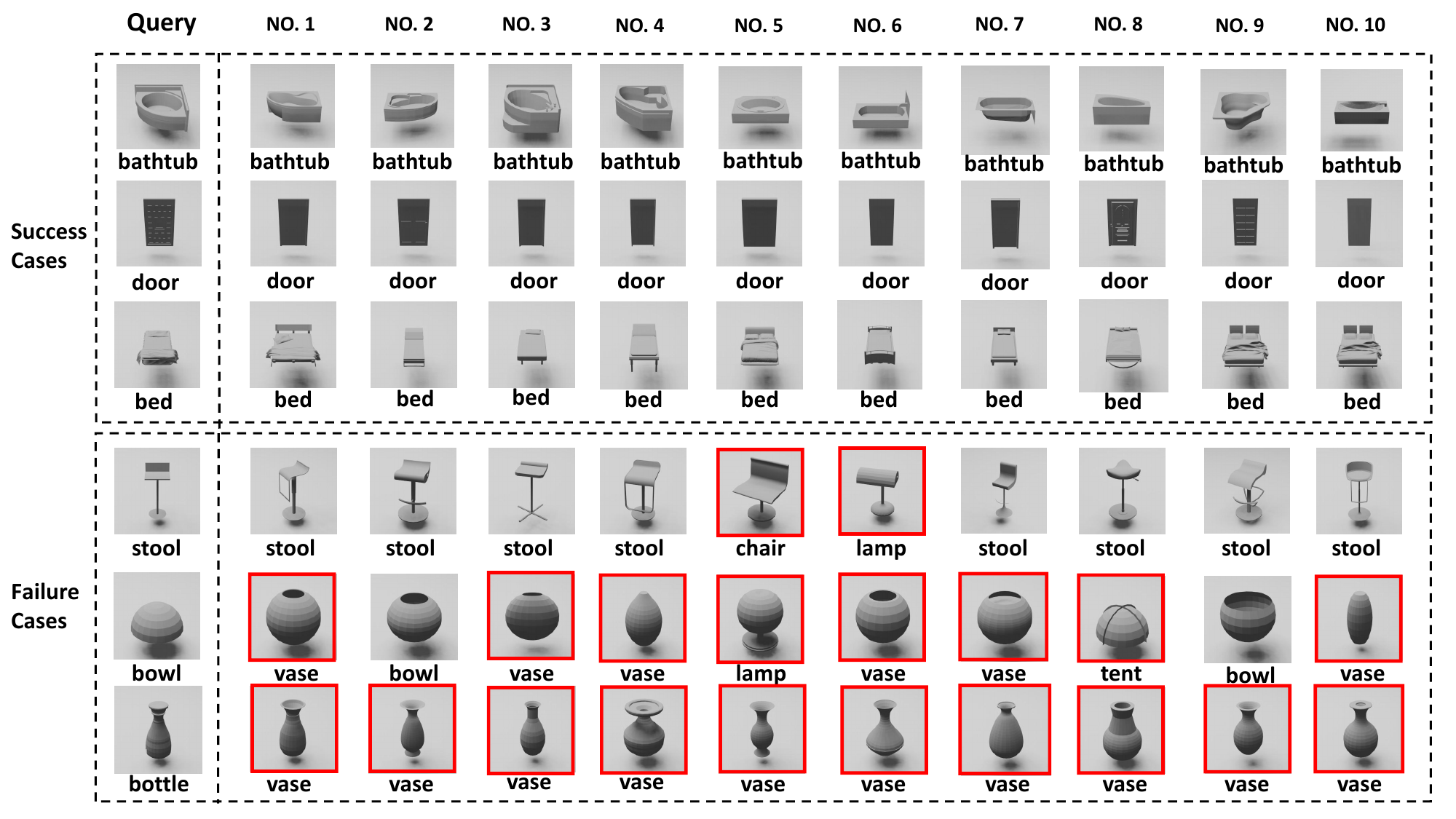}
\caption{\textbf{More retrieval examples of our method on OS-MN40-core.} Incorrect matches are in red boxes.}
\label{fig:more_retrieval}
\vspace{-9pt}
\end{figure}

Figure~\ref{fig:more_retrieval} presents more retrieval examples on OS-MN40-Core.
As shown, DEC faithfully retrieves relevant 3D objects for 3D query objects of common classes such as bathtub, door, and bed. However, certain challenge cases (row 4-6) exist for classes such as stool, bowl, and bottle, leading to failures. For instance, in row 5, a bowl query is incorrectly matched with instances from the vase, despite the subtle differences in their geometric profiles (\emph{e.g.}, aperture width and base structure).
Notably, the top-ranked false positives typically share strong overall geometric similarities with the 3D query objects, which indicates that our model successfully learns high-level shape features but can be confounded by nuanced inter-class variations. 
One potential direction for improvement is to enhance fine-grained class discrimination by modeling these fine-grained geometric variations.

\section{Retrieval on Seen and Unseen Categories}
We further evaluate the performance of our method on both seen and unseen categories.
Following HGM$^{2}$R~\cite{feng2023hypergraph}, we split the ModelNet40 dataset into two subsets: one for seen categories and one for unseen categories. Each subset contains 20 categories, with 80\% of the data used for training (training on seen categories and unseen categories separately) and 20\% reserved for retrieval evaluation (evaluation on seen and unseen categories). The models are trained on the seen categories and evaluated separately on both the seen and unseen category sets.
As shown in Table~\ref{tab:seen_unseen}, for seen categories, our method demonstrates competitive performance on seen categories, relying solely on multi-view images. It achieves a mAP of 94.09 and Recall@100 of 82.63, surpassing other methods in Recall@100. 
While the mAP is slightly below HGM$^2$R (94.10), which requires multi-modalities and test data for training, our method only relies on visual input alone and does not involve test data for training, highlighting our advantages in 3DOR.
For unseen categories, our method outperforms all methods, achieving 86.47 in mAP and 82.00 in Recall@100. These results confirm the model's strong generalization ability, crucial for real-world applications with unseen categories.

\begin{table}[ht]
\small
\setlength{\tabcolsep}{4pt}
\resizebox{1.0\linewidth}{!}{
\centering
\begin{tabular}{lcc|cc}
\toprule
\multirow{2}{*}{Method} 
& \multicolumn{2}{c|}{On Seen Categories} 
& \multicolumn{2}{c}{On Unseen Categories} \\
 & mAP$\uparrow$ & Recall@100$\uparrow$ 
 & mAP$\uparrow$ & Recall@100$\uparrow$     \\
\midrule
TCL~\cite{he2018triplet}    & 93.50 & 82.14 & 73.92 & 71.76 \\
MMJM~\cite{nie2019mmjn}   & 91.99 & 80.78 & 73.07 & 71.38 \\
SDML~\cite{hu2019scalable}   & 88.50 & 78.50 & 74.69 & 72.39 \\
CMCL~\cite{jing2021cross}   & 90.99 & 79.60 & 75.21 & 72.49 \\
MMSAE~\cite{wu2019multi}  & 88.72 & 78.61 & 76.03 & 72.94 \\
MCWSA~\cite{zheng2022multi}  & 85.70 & 76.83 & 72.89 & 70.56 \\
PROSER~\cite{zhou2021learning} & 87.71 & 77.78 & 74.93 & 72.56 \\
InfoNCE~\cite{oord2018representation} & 93.65 & 82.19 & 73.92 & 71.64 \\
HGM$^{2}$R~\cite{feng2023hypergraph} & \textbf{94.10} & 82.47 & 82.23 & 78.21 \\
DAC~\cite{wang2025describe}& 91.12 & 80.46 & 86.27 & 81.76  \\
\cc Our Method & \cc 94.09 & \cc \textbf{82.63} & \cc \textbf{86.47} & \cc \textbf{82.00} \\
\bottomrule
\end{tabular}
}
\caption{\textbf{Separate retrieval results on seen and unseen classes.}}
\label{tab:seen_unseen}
\end{table}

\section{Results on More Realistic Datasets}

\noindent\textbf{OS-Objaverse-core.}
To validate DEC on real-world scenarios, we curate OS-Objaverse-core, a new large-scale open-set 3DOR dataset based on Objaverse-LVIS of Objaverse~\cite{deitke2023objaverse}. Objaverse-LVIS has 1,142 categories in total. 
We filter out categories with fewer than 20 objects, resulting in 607 categories. We split them into training, query, and target subsets, which have 10,092, 1,458, and 15,274 objects, respectively.
The training set comprises 121 seen classes for training, while the query and target contain 486 unseen classes for evaluation. 
Table~\ref{tab:methods_on_OBJ} compares the performance of DEC with state-of-the-art methods on OS-Objaverse-core. 
We also include a zero-shot baseline based on DINOv3~\cite{simeoni2025dinov3} which mean-pools the view features of ViT-L.
As shown, DEC significantly surpasses this zero-shot baseline, increasing the mAP by +3.30\%, demonstrating the effectiveness of DEC for adaptation in this challenging large-scale dataset with highly diverse categories. 
When compared with recent competitors such as CLIP-AdaM~\cite{he2025clip} and DAC~\cite{wang2025describe},  our method consistently achieves remarkable performance gains across all metrics. 
Specifically, we achieve a notable gain of +13.64\% in mAP, +6.60\% in NDCG, and +15.15\% in ANMRR over DAC.
\begin{table}[t]
    \centering
    \tablestyle{10pt}{1.1}
    \begin{tabular}{cccc}\toprule[1pt]
         Method& mAP$\uparrow$&  NDCG$\uparrow$ & ANMRR$\downarrow$ \\
         \midrule
         CLIP-AdaM~\cite{he2025clip}   &  16.97 &  13.71 &  80.39\\
         DAC~\cite{wang2025describe} & 21.89 & 16.42  & 75.86\\
      
         DINOv3~\cite{simeoni2025dinov3} & 32.23 & 21.34 & 66.31\\
         \cc Ours & \cc  \textbf{35.53} & \cc \textbf{23.02} & \cc \textbf{63.71}\\
   \bottomrule[1pt]
 \end{tabular}
    \caption{\textbf{Performance comparison on OS-Objaverse-core.}}
    \label{tab:methods_on_OBJ}
%\vspace{-8pt}
\end{table}

\noindent\textbf{3DFuture.} To further validate DEC’s effectiveness on real-world data, we evaluate it on 3DFuture~\cite{fu20213d}, a large-scale 3D object retrieval benchmark comprising 49 object categories. We split them into training, query, and target subsets, which have 3373, 205, and 11234 objects, respectively. Ten classes are designated as seen and used for training, and the remaining 39 classes are reserved for evaluation.
As shown in Table~\ref{tab:methods_on_3DFuture}, our method significantly outperforms the strong baselines, achieving substantial improvements across all evaluation metrics.
Compared to DAC~\cite{wang2025describe}, we observe a remarkable gain of +6.30\% in mAP, +4.23\% in NDCG, and a -5.44\% reduction in ANMRR, confirming the efficacy of our design in handling the high category diversity and scale of the 3DFuture dataset. 
Compared with recent DINOv3~\cite{simeoni2025dinov3}, we yield gains of +4.68\% in mAP, +1.99\% in NDCG, and -3.58\% in ANMRR.
These results demonstrate that our method offers a scalable and effective solution for real-world 3D object retrieval scenarios in open-set conditions.
\begin{table}[t]
    \centering
    \tablestyle{10pt}{1.1}
    \resizebox{0.95\linewidth}{!}{%
    \begin{tabular}{cccc}\toprule[1pt]
         Method& mAP$\uparrow$&  NDCG$\uparrow$ & ANMRR$\downarrow$ \\
         \midrule
         DAC~\cite{wang2025describe} & 29.40 & 43.06  & 68.40 \\
         DINOv3~\cite{simeoni2025dinov3} & 31.02 & 45.30 & 66.54\\
         \cc \textbf{Ours} & \cc  \textbf{35.70} & \cc \textbf{47.29} & \cc \textbf{62.96}\\
   \bottomrule[1pt]
 \end{tabular}}
    \caption{\textbf{Performance comparison on 3DFuture.}}
    \label{tab:methods_on_3DFuture}
\end{table}

\noindent\textbf{ScanObjectNN.} Our DEC is also \textit{compatible with point} \textit{clouds} by projecting it online into depth maps.
For the experiment, 10 depth maps are projected for each point cloud online following~\cite{zhu2023pointclip}.
We experiment on noisy real-world ScanObjectNN. We achieve 30.34\% mAP, outperforming recent CLIP-Adam~\cite{he2025clip} greatly. The zero-shot baseline attains only 23.77\% mAP.

\begin{table}[t]
    \centering
    \tablestyle{10pt}{1.1}
    \begin{tabular}{cccc}\toprule[1pt]
         Method& mAP \\
         \midrule
         \textit{Base.} &  23.77 \\
         CLIP-Adam~\cite{he2025clip} & 24.18 \\
         \cc \textbf{Ours} & \cc  \textbf{30.34}\\
   \bottomrule[1pt]
 \end{tabular}
    \caption{\textbf{Performance comparison on ScanObjectNN.}}
    \label{tab:methods_on_scan}
%\vspace{-16pt}
\end{table}

\begin{table}[t]
    \centering
    \tablestyle{10pt}{1.1}
    \begin{tabular}{cccc}\toprule[1pt]
         Method& Top-1 Accuracy \\
         \midrule
         PointCLIP V2~\cite{zhu2023pointclip} & 89.55 \\
        
         PointNet++~\cite{qi2017pointnet++} & 90.7 \\
         \cc \textbf{Ours} & \cc  \textbf{91.49}\\
   \bottomrule[1pt]
 \end{tabular}
    \caption{\textbf{Few-shot learning results on ModelNet40.} We report the 16 shot classification accuracy (\%).}
    \label{tab:methods_on_ModelNet40}
%\vspace{-16pt}
\end{table}

\section{Extending to Few-shot Classification}
\noindent\textbf{3D classification.} Beyond open-set 3D object retrieval, we further evaluate DEC on diverse downstream tasks to assess its general-purpose representation capability. We conduct a 16-shot 3D classification on ModelNet40.
As shown in Table~\ref{tab:methods_on_ModelNet40}, we surpass PointCLIP V2~\cite{zhu2023pointclip} greatly on Top-1 accuracy, even surpassing the fully supervised PointNet++~\cite{qi2017pointnet++} without \textit{normal}.


\end{document}